\documentclass[10pt,twocolumn,letterpaper]{article}

\usepackage[final]{cvpr}      

\usepackage{makecell}
\usepackage{placeins}
\usepackage{amsmath}
\usepackage{amssymb}
\usepackage{algorithm}
\usepackage{algpseudocode}
\usepackage{tabularx}
\usepackage{comment}
\usepackage{multirow} 
\usepackage{longtable}
\usepackage{booktabs}
\usepackage{amssymb}
\usepackage{amsmath}
\usepackage{graphicx} 
\usepackage{float}

\definecolor{cvprblue}{rgb}{0.21,0.49,0.74}
\usepackage[pagebackref,breaklinks,colorlinks,allcolors=cvprblue]{hyperref}

\def\paperID{33248} 
\def\confName{CVPR}
\def\confYear{2026}

\title{Spatial Transcriptomics as Images for Large-Scale Pretraining}

\author{
Yishun Zhu$^{1,2}$\thanks{Both authors contributed equally to this research.}\quad
Jiaxin Qi$^{1,2}$\footnotemark[1]\quad
Jian Wang$^{1}$\quad
Yuhua Zheng$^{2}$\thanks{Corresponding author.}\quad
Jianqiang Huang$^{1}$\footnotemark[2]\\[0.5em]
$^{1}$Computer Network Information Center, Chinese Academy of Sciences, Beijing, China\\
$^{2}$Hangzhou Institute for Advanced Study, University of the Chinese Academy of Sciences, Zhejiang, China\\[0.3em]
{\tt\small zhuyishun25@mails.ucas.ac.cn, zhengyuhua@ucas.ac.cn, \{jxqi, wangjian, jqhuang\}@cnic.cn}\\[0.3em]
{\tt\small Code: \url{https://github.com/Lan48/treat-sc-img}}
}

\begin{document}
\maketitle
\begin{abstract}

Spatial Transcriptomics (ST) profiles thousands of gene expression values at discrete spots with precise coordinates on tissue sections, preserving spatial context essential for clinical and pathological studies.
With rising sequencing throughput and advancing platforms, the expanding data volumes motivate large-scale ST pretraining.
However, the fundamental unit for pretraining, i.e., what constitutes a single training sample, is not well defined. Existing choices fall into two camps: (1) treating each spot as an independent sample, which discards spatial dependencies and collapses ST into single-cell transcriptomics; and (2) treating an entire slice as a single sample, which produces prohibitively large inputs and drastically fewer training examples, undermining effective pretraining.
To address this gap, we propose treating spatial transcriptomics as croppable images. Specifically, we define a multi-channel image representation with fixed spatial size by cropping patches from raw slices, thereby preserving spatial context while substantially increasing the number of training samples. Along the channel dimension, we define gene subset selection rules to control input dimensionality and improve pretraining stability.
Extensive experiments show that the proposed image-like dataset construction for spatial transcriptomics pretraining consistently improves downstream performance, outperforming conventional pretraining schemes. Ablation studies verify that both spatial patching and channel design are necessary, establishing a unified, practical paradigm for organizing ST data and enabling large-scale pretraining.


\end{abstract}    
\begin{figure}[!t]
  \centering
  \includegraphics[width=\linewidth]{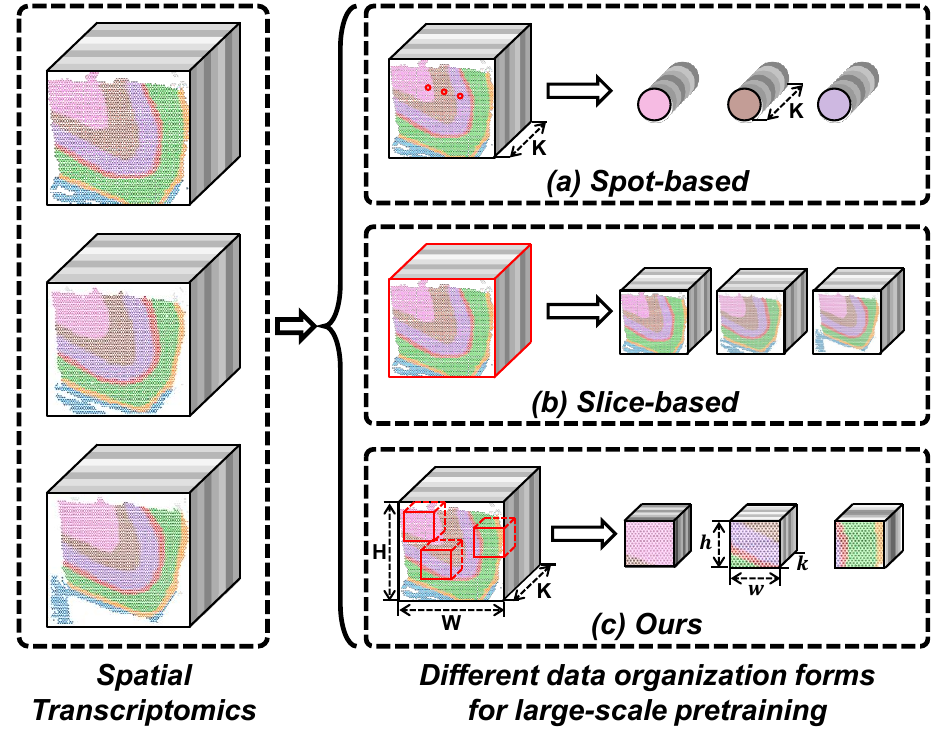}
  \caption{
Illustration of data organization strategies for large-scale spatial transcriptomics pretraining.
(a) Spot-based method: each spot is treated as an independent \(K\)-dimensional gene-expression vector, discarding spatial relationships.
(b) Slice-based method: each whole tissue slice is treated as a single training sample, leading to very large inputs and few examples.
(c) Ours: spatial transcriptomics is treated as croppable multi-channel images by extracting fixed-size spatial patches (\(h \times w\)) with selected gene subsets along the channel dimension (\(k\)), preserving local spatial context while greatly increasing the number of training samples.
  }
  \label{fig:teaser}
\end{figure}
\section{Introduction}
\label{sec:intro}

Spatial transcriptomics (ST)~\cite{Marx_2021} has emerged as a transformative technology for profiling gene expression directly on intact tissue sections while preserving their spatial organization~\cite{Burgess_2019, Asp_2020}.
The resulting data can be regarded as a collection of discrete spots, each associated with a high-dimensional vector of gene expression values and its precise 2D coordinate.
This spatially resolved representation enables quantitative analysis of tissue architecture, cellular neighborhoods, and microenvironments (e.g., brain regions) that remain inaccessible to bulk and single-cell transcriptomics, which lack spatial context~\cite{Chung2024AccurateSG,Miller2021MultiSD,YANG2024109966,Systematic_comparison}. 
Thanks to advances in sequencing technologies and platforms such as 10x Genomics Visium~\cite{Visium}, an increasing number of ST datasets are being released across diverse organs, species, and disease conditions, enabling large-scale pretraining and the learning of generic spatially aware representations for downstream spatial biology tasks~\cite{SpatialScope}.

However, when scaling spatial transcriptomics to large-scale pretraining, the fundamental training unit, i.e., what should constitute a single training sample, remains unclear. Existing designs fall into two camps, as illustrated in Figure~\ref{fig:teaser}, (a) Spot-based methods treat each spot as an independent sample, keep gene expression dimension $K$, and learn to predict masked gene expression at the spot level.
The main advantage of this formulation is that it substantially increases the number of training examples. For example, a typical ST slice with a $50 \times 50$ grid already yields $2{,}500$ spot-level samples. However, this strategy inevitably discards most spatial structure: even if coordinates or local neighborhoods are appended as features for the spot~\cite{SCGPTSpatial}, the model effectively sees a bag of isolated spots, collapsing spatial transcriptomics into single-cell–like transcriptomics and losing the rich spatial context.
In contrast, (b) slice-based methods regard an entire tissue slice as a single sample, preserving global spatial organization and long-range dependencies. However, the resulting input becomes extremely large,
leading to prohibitive computation and making pretraining ineffective. Meanwhile, the number of available slices remains limited, providing too few training examples to fully exploit large-scale pretraining. Therefore, neither spot-based nor slice-based formulations offer a satisfactory basic unit for pretraining ST foundation models.


Motivated by these limitations, we propose treating spatial transcriptomics as croppable multi-channel images, bridging the limited spatial context of spot-based sampling and the prohibitive computation of slice-level methods. Specifically, we view each ST slice as a 2D grid indexed by spatial coordinates, where every spot corresponds to a pixel and its gene expression profile forms a high-dimensional channel vector. Rather than operating on isolated spots or entire slices, as shown in Figure~\ref{fig:teaser}(c), we sample fixed-size $h \times w$ windows from this grid, producing image-like patches that preserve local spatial organization while substantially increasing the number of training samples. Along the channel dimension, we further introduce an importance-aware gene subset selection strategy, where in each window, we stochastically sample a subset of highly variable genes, consistent with biological notions of gene importance, yielding compact yet informative channels that control input dimensionality and stabilize pretraining. Our image-like formulation strikes a practical balance between spatial fidelity, computational efficiency, and sample scales, providing a scalable basic unit for large-scale ST pretraining.

We conduct extensive experiments on ST downstream tasks, such as spatial domain detection, across multiple datasets. Models pre-trained on our patch-based ST formulation consistently outperform counterparts trained with conventional schemes, demonstrating that treating ST as images yields more scalable and data-efficient pretraining. Furthermore, by standardizing the sample construction, gene-channel configuration, and evaluation protocol, our framework provides a unified and reproducible recipe for building ST pretraining datasets and benchmarks, offering a practical foundation for future large-scale ST representation learning and downstream applications.

Our contributions are summarized as follows:
\begin{itemize}
\item We systematically analyze the sample design of ST pretraining, highlighting the limitations of existing spot-based and slice-based formulations and clarifying their respective strengths and weaknesses in terms of spatial fidelity and sample scale.
\item We propose a patch-based formulation that crops fixed-size multi-channel patches from ST slices, coupled with an importance-aware gene subset selection along the channel dimension. Our design preserves local spatial context while controlling input dimensionality, leading to stable and scalable ST pretraining.
\item We perform extensive experiments and ablation studies to demonstrate the effectiveness of the proposed data construction across multiple downstream datasets. Finally, we establish a standardized, reproducible paradigm for organizing ST data into pretraining datasets and benchmarks, offering a solid foundation for future large-scale ST pretraining studies.
\end{itemize}

\section{Related Works}
\label{sec:related works}





\noindent\textbf{Spatial Transcriptomics.} Spatial Transcriptomics (ST) is a cutting-edge technology that enables high-throughput detection and analysis of gene expression patterns within the spatial context of tissue samples. The key spatial transcriptomics sequencing technologies currently include Visium~\cite{Visium}, VisiumHD~\cite{Visium}, Xenium~\cite{Xenium}, and MERFISH~\cite{MERFISH,MERFISH2}. By integrating high-throughput sequencing or high-precision imaging methods, these technologies can quantitatively map RNA molecules across tissue sections. By revealing the spatial relationships between tissue structures and gene expression, these techniques play a pivotal role in studying embryonic development, tumor microenvironments, and gene-target localization, opening new frontiers in the study of human tissue development, disease progression, and clinical diagnosis~\cite{He2020IntegratingSG,doi:10.1126/science.aav9776,2019A}.

\noindent\textbf{Pretraining for Spatial Transcriptomics.} With the advent of large-scale ST corpora, two primary pretraining paradigms have emerged: \emph{spot-based} and \emph{slice-based}.

\noindent\textit{(1) Spot-based pretraining.} Several recent foundation models focus on the spot level, where each sequencing unit (spot) or a small cellular neighborhood is modeled as a sample. For example, scGPT-spatial~\cite{SCGPTSpatial} pretrains a single-cell foundation model on $\sim$30M spatial profiles with a spatially aware sampling objective to capture local co-localization patterns, achieving strong performance in cell-type deconvolution and missing-gene imputation. Meanwhile, ST-Align~\cite{Lin2024STAlignAM} addresses multimodal image–gene alignment by pretraining on $\sim$1.3M spot–niche pairs with a three-target alignment strategy (spot, niche, slide), thereby bridging pathology imaging with gene-expression features and yielding superior zero-/few-shot transfer.
And some method (STAGATE~\cite{STAGATE}, GraphST~\cite{GraphST}, STitch3D~\cite{STitch3D}) organize spots into graphs to learn the spatial context.

\noindent\textit{(2) Slice-based pretraining.} At the slice level, modeling spans entire tissue sections or multiple adjacent slices, capturing richer spatial context and multi-scale structure. For instance, SToFM~\cite{stofm} proposes a multi-scale foundation model that extracts macro-scale tissue morphology, micro-scale cellular microenvironment, and gene-scale expression features via an SE(2)Transformer~\cite{SE(2)} and the curated pretraining corpus SToCorpus-88M~\cite{stofm}. Another example, FmH2ST~\cite{FmH2ST}, adopts a dual-branch architecture integrating slice-level image context and spot-level detail to address inter-slice heterogeneity and intra-slice complexity for spatial gene-expression prediction.%

Despite recent progress, current ST pretraining remains constrained by two extremes: spot-based methods treat each spot independently, discarding spatial dependencies and effectively collapsing ST into single-cell pretraining; slice-based methods model entire sections as single instances, yielding prohibitively large inputs and drastically fewer training examples. We therefore introduce a large-scale pretraining framework that moves beyond spot- or slice-centric formulations, enabling image-like modeling that unifies spatial context across slices and samples.

\section{Method}
\label{sec:formatting}

\subsection{Preliminaries}
\label{sec:preliminaries}

\noindent\textbf{Spatial Transcriptomics Data}. Different from single-cell transcriptomic data, whose samples only contain the gene expressions aligned to a shared gene set, spatial transcriptomics (ST) augments each expression profile with a 2D coordinate on the tissue slice. Let \(\mathcal{G}\!=\!\{g_1,g_2,\dots,g_K\}\) denote the gene set and \(\bm{v}\!=\!\{v_1,v_2,\ldots,v_K\}\) denote the corresponding expression values.
For a slice with \(H\!\times\! W\) grid with the spot index set \(\Omega\!=\!\{1,\ldots,H\}\!\times\!\{1,\ldots,W\}\), we represent the slice as:
\begin{equation}
\label{eq:stdata}
\bm{s}=\{(\bm{v}_{\bm{c}},\bm{c})\}_{\bm{c}\in\Omega},\qquad
\bm{c}=(x,y)\in\mathbb{Z}^{2}.
\end{equation}
Each spot is thus characterized by its gene expression and spatial location. The whole dataset for ST could be represented as a collection of slices, $\mathcal{D}\!=\!\{\bm{s}_i\}_{i=1}^N$. ST pretraining aims to learn an encoder \(f_\theta\) that captures joint spatial-molecular structure, yielding spot-based or slice-based representations for downstream tasks.

\noindent\textbf{Pretraining Frameworks}. We review three self-supervised pretraining strategies in spatial transcriptomics, categorized by the spatial scale of the context they incorporate.

\medskip
\noindent\textit{Spot-based Pretraining}.
This is the finest granularity, representation learning at individual spots (treating each spot as a training sample). Given a spot \(i\) with expression vector \(\bm{v}_i\), we sample a mask set \(\mathcal{M}_i\subseteq\{1,\ldots,K\}\) and form a masked input value $\tilde{\bm{v}}_i$. Given the model with the encoder $\phi_\theta$, the objective for spot-based pretraining is to predict the masked expression values at the spot level:
\begin{equation}
\label{eq:spotloss_unified}
\mathcal{L}_{\mathrm{spot}}
= \frac{1}{N_{\text{sp}}|\mathcal{M}_i|}\sum_{i=1}^{N_{\text{sp}}}
\sum_{k\in\mathcal{M}_i}
\big(v_{i,k}-\bm{e}_{\bm{g}_k}^T\phi_\theta(\tilde{\bm{v}}_i)\big)^2,
\end{equation}
where $N_{\text{sp}}$ is the total number of spots, $\bm{e}_k$ is the embedding of gene $k$, and $\bm{e}_{\bm{g}_k}^T\phi_\theta(\tilde{\bm{v}}_i)$ is the predicted expression value for the $k$-th masked gene, \ie, $\hat{\bm{v}}_{i,k}$.
Across a dataset of \(N\) slices, the total number of spot-level training examples is \(N_{\text{sp}}= H\!\times\! W\!\times\! N\), thus spot-level pretraining yields abundant samples but discards positional information, degenerate spatial transcriptomic pretraining.

\medskip
\noindent\textit{Multi-spot Pretraining}.
To capture short-range spatial correlation, this strategy (e.g., scGPT--Spatial~\cite{SCGPTSpatial}) augments each spot with a neighborhood-aware context aggregated from adjacent spots. Then, the embedding of a spot $i$ could be represented by its neighbors to improve the spatial awareness.
Let \(\mathcal{N}(i)\) denote the set of spatial neighborhood for spot \(i\), and let \(\bm{e}_j\!=\!\phi(\bm{v}_j)\) denote the embedding of a neighbor \(j\!\in\!\mathcal{N}(i)\), and we aggregate the neighbor embeddings via weights \(\alpha_{ij}\) to form neighbor enhanced spot embedding for $i$, \ie, 
\(\overline{\bm{e}}_i=\sum_{j\in\mathcal{N}(i)}\alpha_{ij}\bm{e}_j\).
Then, the pretraining loss for masked genes of spot $i$ conditioning on its neighbors could be written as:
\begin{equation}
\mathcal{L}_{\mathrm{mspot}}
\!=\! \frac{1}{N_{\text{sp}}|\mathcal{M}_i|}\sum_{i=1}^{N_{\text{sp}}}
\sum_{k\in\mathcal{M}_i}
\big(v_{i,k}-\bm{e}_{\bm{g}_k}^T\overline{\bm{e}}_i\big)^2,
\label{eq:multispotloss_unified}
\end{equation}
where other notations follows Eq.~\eqref{eq:spotloss_unified}.
Although \(\overline{\bm{e}}_i\) injects local spatial context and partially remedies the spatial-agnostic nature of spot-based methods, its receptive field is inherently short-range and weakly tied to absolute geometry, leaving macro-scale structures under-exploited and limiting spatial awareness beyond immediate neighborhoods.

\medskip
\noindent\textit{Slice-based Pretraining}.
This strategy (e.g., SToFM~\cite{stofm}) injects global slice-level context to capture macro-scale tissue organization. Let the spots of a slice be partitioned into $k$ macro-domains, $\{\mathcal{C}_1,\ldots,\mathcal{C}_k\}$, and let $\mathcal{C}(i)$ denote the domain containing spot $i$. The embedding $\bm{e}_{\mathcal{C}(i)}$ is constructed by aggregating the encoded spot features $\{\bm{e}_j\}_{j \in \mathcal{C}(i)}$ from all spots belonging to the macro-domain $\mathcal{C}(i)$ of spot $i$. 
The global-context-aware embedding is formed by fusing the spot embedding $\bm{e}_i$ and its macro-domain context: $\bm{e}_i^{\mathrm{s}} \!=\! f(\bm{e}_i, \bm{e}_{\mathcal{C}(i)})$.
The pretraining loss conditioning on this slice-aware context for the masked gene $k$ 
could be written as:

\begin{equation}
\mathcal{L}_{\mathrm{slice}}
\!=\! \frac{1}{N_{\text{sl}}|\mathcal{M}_i|}\sum_{i=1}^{N_{\text{sl}}}
\sum_{k\in\mathcal{M}_i}
\big(v_{i,k}-\bm{e}_{\bm{g}_k}^T\bm{e}_i^{\mathrm{s}}\big)^2,
\label{eq:slice}
\end{equation}
where $N_\text{sl}$ is the number of macro-domains, and when $k\!=\!1$, $N_\text{sl}\!=\!N$.
This objective acts as a global regularizer by aligning spot embeddings with slice-level tissue structure. By explicitly modeling spatial context, slice-based pretraining achieves global spatial awareness beyond local neighborhoods.
However, constructing and fusing slice-scale context substantially increases input size, and macro-domain delineation lacks a unified standard across datasets. In the extreme where the entire slice is treated as the context unit, the effective number of training samples collapses to the number of slices, which is typically small, hindering large-scale ST pretraining.

\begin{table*}[!t]
    \centering
\caption{Comparison of different methods for spatial domain detection using MLP classification on downstream task datasets. Test accuracy (Acc) and adjusted Rand index (ARI) are reported. \textit{Raw}: using original gene expression directly. \textit{scGPT}: using scGPT embeddings. \textit{scGPT-sp}: scGPT-spatial. \textit{Ours(k)}: our method with $k$ which refers to the spatial dimensions (width and height) of pretraining samples.}
\label{tab1}
\setlength{\tabcolsep}{4.75pt}
\setlength{\aboverulesep}{0pt} 
\setlength{\belowrulesep}{0pt} 

    \begin{tabularx}{\textwidth}{c|cccccccccccccc}\toprule
         \multirow{2}{*}{Method}&  \multicolumn{2}{c}{DLPFC}&  \multicolumn{2}{c}{LNA}&  \multicolumn{2}{c}{LND}&\multicolumn{2}{c}{Col}&\multicolumn{2}{c}{BrC}&\multicolumn{2}{c}{Ton}&\multicolumn{2}{c}{Avg}\\\cmidrule{2-15}
 & Acc& ARI & Acc& ARI & Acc& ARI & Acc& ARI & Acc& ARI & Acc& ARI & Acc& ARI\\\midrule
         raw&  0.571  & 0.305&  0.469  &0.146&  0.430 &0.109&  0.489 & 0.254&  0.645  &0.550&  \textbf{0.729} &\textbf{0.349}&0.555 &0.285\\
         scGPT &  0.377  & 0.046&  0.489  &0.144&  0.433  &0.115&  0.207  &0.013&  0.205  & 0.110&  0.538  &0.057&0.375 &0.081\\
         scGPT-sp&  0.678  &0.437&  0.570  &0.299&  0.573  &0.319&  0.423  &0.160&  0.716  &0.589&  0.660  & 0.247&0.603 &0.342\\\midrule
         ours(8)&  0.652&0.370&  0.665  &0.390&  0.609  &0.349&  0.383  &0.143&  0.764  &0.616&  0.659  &0.198
&0.622 &0.344\\
         ours(16)&  \textbf{0.753} & \textbf{0.580}&  \textbf{0.670} &\textbf{0.393}&  \textbf{0.635} &\textbf{0.381}&  0.467 &0.234&  \textbf{0.773} &\textbf{0.709}&  0.674 &0.268
&\textbf{0.662} &\textbf{0.428}\\
         ours(24)&  0.720  &0.537&  0.663  &0.390&  0.580  & 0.289&  \textbf{0.491} &\textbf{0.258}&  0.712  &0.629&  0.669  &0.235
&0.639 &0.390\\
         ours(30)& 0.667  &0.434& 0.631  &0.348& 0.586  &0.284& 0.482 &0.247& 0.693  &0.604& 0.667  &0.241
&0.621 &0.360\\ \bottomrule
    \end{tabularx}
\end{table*}


\begin{table}[!t]
\centering
\caption{Accuracy comparison with SToFM for spatial domain detection using MLP classification on downstream task datasets.}
{\setlength{\tabcolsep}{4pt}
 \renewcommand{\arraystretch}{0.98}
 \small 
\begin{tabular}{l!{\vrule width 1pt} *{7}{c}}
\hline
Method & DLPFC & LNA & LND & Col & BrC & Ton & Avg \\
\hline
ours(16) & 0.753 & 0.670 & 0.635 & 0.467 & 0.773 & 0.674 & 0.662 \\
sToFM    & 0.645 & 0.602 & 0.555 & 0.461 & 0.443 & 0.658 & 0.542\\
\hline
\end{tabular}
} 
\label{tab:VSsToFM}
\end{table}

\subsection{Our Method}
\label{sec:proposed}
Prior spot-based and slice-based pretraining frameworks either ignore spatial context or suffer from poor scalability. Thus, we propose to treat spatial transcriptomics as images and introduce a fixed-size patch-level ST pretraining framework. Specifically, we treat each ST slice as a croppable multi-channel image and sample unified fixed-size windows of shape \(h\!\times\! w\) as training units. These units preserve local topology while yielding far more training examples than slice-based sampling and avoiding the loss of spatial context of spot-based methods. To bound input dimensionality without sacrificing biological information, each patch adopts an importance-aware gene subset selection, where we stochastically select \(m\) highly variable genes within the window.
This image-like formulation bridges spot modeling and slice context, providing a scalable and generalizable basic unit for large-scale ST pretraining.

\smallskip
\noindent\textbf{Sample Construction}. We construct a unified patch-based basic pretraining unit for ST in three steps.

\begin{figure}[t]
\centering
\includegraphics[width=\linewidth]{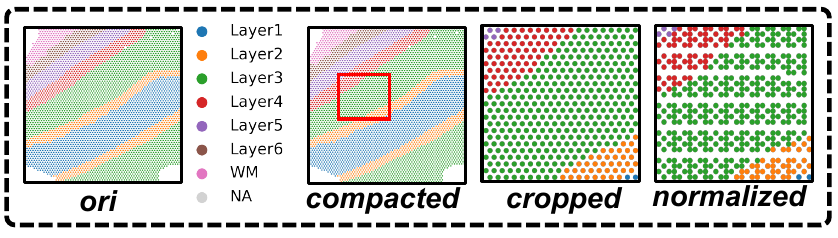}
\caption{Sample construction — ori (original), compacted (scaled), cropped (fixed-window), normalized (grid-mapped). Visualization: spot size 100 for ori, 1 for others.}
\label{fig-samples}
\end{figure}

\begin{algorithm}[t]
\caption{Our Sample Construction Framework}
\label{alg:sample_construction}
\begin{algorithmic}[1]
    \State \textbf{Input:} ST dataset $\mathcal{D}\!=\!\{\bm{s}_i\}_{i=1}^N$, hyperparameters $h,w,m$, and number of samples per slice $n_s$
    \State \textbf{Output:} Patch-based dataset $\mathcal{D}_{\mathrm{ours}}$
    \State $\mathcal{D}_{\mathrm{ours}} \gets \emptyset$
    \For{$i = 1$ to $N$} \Comment{process each slice $\bm{s}_i$}
        \State $\Omega^\star \gets \{\bm{c} \mid (\bm{v}_{\bm{c}}, \bm{c}) \in \bm{s}_i\}$ \Comment{remove empty spot}
        \State $X \gets \mathrm{sorted}(\mathrm{unique}(\{x \mid (x, y) \in \Omega^\star\}))$
        \State $Y \gets \mathrm{sorted}(\mathrm{unique}(\{y \mid (x, y) \in \Omega^\star\}))$
        \State $W' \gets |X|$; \; $H' \gets |Y|$
        \State Initialize $\bm{s}_{\text{temp}} \in \mathbb{R}^{H' \times W' \times K}$ with zeros
        \ForAll{$(\bm{v}_{(x,y)}, (x,y)) \in \bm{s}_i$}
            \State Map $(x,y)$ to indices $(c_x,c_y)$ by Eq.~\eqref{eq:rank_map}
            \State $\bm{s}_{\text{temp}}[c_x, c_y, :] \gets \bm{v}_{(x,y)}$
        \EndFor
        \For{$r = 1$ to $n_{\mathrm{win}}$} \Comment{randomly sample patches}
            \State Sample $\bm{t}$ on $\bm{s}_{\text{temp}}$ by Eq.~\eqref{eq:rand_window}
            \State Compute per-gene variance $\sigma_g^2$ in $\bm{t}$
            \State $\mathcal{G}_{\mathrm{sel}} \sim \mathrm{Sample}_{m}(\{1, \dots, \ell\}; \Pr(g))$ \Comment{Eq.~\eqref{eq:genesel}}
            \State $\bm{t}' \gets \bm{t}[:, :, \mathcal{G}_{\mathrm{sel}}]$
            \State $\mathcal{D}_{\mathrm{ours}} \gets \mathcal{D}_{\mathrm{ours}} \cup \{\bm{t}'\}$
        \EndFor
    \EndFor
    \State \Return $\mathcal{D}_{\mathrm{ours}}$
\end{algorithmic}
\end{algorithm}

\smallskip
\noindent\textit{(1) Coordinate Normalization.}
Many ST platforms yield staggered layouts with structured holes inside the rectangular index set \(\Omega\). To obtain a tight sampling lattice without fabricating measurements, we compact only the observed positions \(\Omega^\star\subseteq\Omega\) while preserving local adjacency. Concretely, after removing unobserved sites, we build an \(H'\times W'\) grid from the counts of distinct horizontal/vertical levels. 
Let \(X\!=\!\{x_1\!<\!\cdots\!<\!x_{W'}\}\) and \(Y\!=\!\{y_1\!<\!\cdots\!<\!y_{H'}\}\) be the sorted unique absolute coordinates of the remaining spots. 
Each absolute coordinate \((x,y)\) is mapped by relative order to discrete grid indices and a normalized position:
\begin{equation}
\label{eq:rank_map}
\begin{aligned}
&c_x=\mathrm{rank}_X(x),\;
c_y=\mathrm{rank}_Y(y).
\end{aligned}
\end{equation}
This rank-based compaction preserves local ordering without imputation, yields tight lattice coordinates \(\bm{c}=(c_x,c_y)\) for subsequent processing.

\smallskip
\noindent\textit{(2) Random Window Cropping.}
On the compact grid \(\Omega^\star\) of size \(H'\times W'\), we randomly crop an \(h\times w\) window as the training unit.
Let \(\bm{o}=(o_x,o_y)\) denote the top-left corner drawn uniformly from \(\{1,\ldots,W'-w+1\}\times\{1,\ldots,H'-h+1\}\).
We rasterize the window into a dense tensor \(\bm{t}\in\mathbb{R}^{h\times w\times \ell}\) (\(\ell\le K\)) as:
\begin{equation}
\label{eq:rand_window}
\begin{aligned}
\bm{o} \sim \mathrm{Unif}\!\big(\{1{:}W'\!-\!w\!+\!1\}\!\times\!\{1{:}H'\!-\!h\!+\!1\}\big),\\
\bm{t}[v,u,:] =\bm{v}_{\alpha_{u,v}}, \; (u,v)\in\{1{:}w\}\times\{1{:}h\}.
\end{aligned}
\end{equation}
where \(\alpha_{u,v}\!=\!(o_x\!+\!u\!-\!1,\;o_y\!+\!v\!-\!1)\) is the absolute grid index of the \((u,v)\)-th spot in $\bm{t}$, and \(\sim\mathrm{Unif}(\mathcal{S})\) denotes sampling from the discrete uniform distribution over the finite set \(\mathcal{S}\). Importantly, each \(\bm{t}\) is treated as an independent local context for pretraining.

\smallskip
\noindent\textit{(3) Channel (Gene) Selection.}
To map the window’s variable gene dimension \(\ell\) to a fixed channel size \(m\), we compute per-gene variance \(\sigma_g^2\) in the window and sample \(m\) unique genes by variance-weighted selection:
\begin{equation}
\label{eq:genesel}
\mathcal{G}_{\mathrm{sel}}
\sim \mathrm{Sample}_{m}\!\Big(\{1,\ldots,\ell\};\; \Pr(g)\propto \sigma_g^2+\varepsilon\Big).
\end{equation}
Here \(\mathrm{Sample}_{m}(\cdot;\Pr)\) denotes weighted sampling without replacement of \(m\) indices, and \(\varepsilon>0\) is for smoothing.
The training unit is then written as \(\bm{t}^{'}=\bm{t}[:,:,\,\mathcal{G}_{\mathrm{sel}}]\in\mathbb{R}^{h\times w\times m}\). We summarize the complete process of sample construction in algorithm \ref{alg:sample_construction}.
Figure \ref{fig-samples} provides an example of our sample construction process, demonstrating our approach to maintaining the relative spatial layout of spots while generating samples.

Putting the steps together, each cropped region becomes an image-like patch with fixed spatial size and channel count. Rank-based compaction preserves neighborhood structure without inventing measurements, random windowing converts every slice into many locally coherent training samples, and variance-weighted gene subsampling stabilizes the channel dimension while emphasizing informative signals and controls the input size.

\medskip
\noindent\textbf{Pretraining Objective}.
Given a cropped training unit $\bm{t}'\!\in\!\mathbb{R}^{h\times w\times m}$, sample masked indices
$\mathcal{M}\!\subseteq\!\{1{:}h\}\!\times\!\{1{:}w\}\!\times\!\mathcal{G}_{\mathrm{sel}}$
and form a masked input $\tilde{\bm{t}}'$. The encoder $\phi_\theta$ outputs
site features $\phi_\theta(\tilde{\bm{t}}')[u,v]$, which are projected to the
$k$-th gene channel via $\bm{e}_{\bm{g}_k}^{\!\top}\!$. The masked
reconstruction objective is:
\begin{equation}
\label{eq:patch_mask}
\mathcal{L}_{\mathrm{ours}}
= \frac{1}{N'|\mathcal{M}|}\sum_{i=1}^{N'}
\sum_{(u,v,k)\in\mathcal{M}}
\Big(t'_{u,v,k}-\bm{e}_{\bm{g}_k}^{\!\top}\phi_\theta(\tilde{\bm{t}}')[u,v]\Big)^2,
\end{equation}
where \(N'\) is the total number of samples, $\phi_\theta(\tilde{\bm{t}}')[u,v]$ denotes the representation at site $(u,v)$, and $\bm{e}_{\bm{g}_k}$ is the embedding of gene $k$. The patch-level design preserves local topology while injecting slice context within each window, yields substantially more pretraining samples than slice-level schemes, and is compatible with vision-style self-supervision. Trade-offs include higher memory/compute for large $(h,w,m)$ and fewer units than purely spot-level training, mitigated by random cropping and stochastic channel subsampling.

\begin{table*}
    \centering
\caption{Comparison of different methods for spatial domain detection using k-nearest neighbors classification on downstream task datasets. Test accuracy (Acc) and adjusted Rand index (ARI) are reported. \textit{Raw}: using original gene expression directly. \textit{scGPT}: using scGPT embeddings. \textit{scGPT-sp}: scGPT-spatial. \textit{Ours k}: our method with $k$ which refers to the spatial dimensions (width and height) of pretraining samples.}
\label{tab2}
\setlength{\tabcolsep}{4.75pt}
\setlength{\aboverulesep}{0pt} 
\setlength{\belowrulesep}{0pt} 
    \begin{tabularx}{\textwidth}{c|cccccccccccccc}\toprule
         \multirow{2}{*}{Method}&  \multicolumn{2}{c}{DLPFC}&  \multicolumn{2}{c}{LNA}&  \multicolumn{2}{c}{LND}&\multicolumn{2}{c}{Col}&\multicolumn{2}{c}{BrC}&\multicolumn{2}{c}{Ton}&\multicolumn{2}{c}{Avg}\\\cmidrule{2-15}
& Acc& ARI & Acc& ARI & Acc& ARI & Acc& ARI & Acc& ARI & Acc& ARI & Acc& ARI\\\midrule
         raw&  0.441&0.142&  0.262&0.077&  0.433&0.040&  0.247&0.021&  0.596&0.433&  0.704&0.305&0.447 &0.170\\
         scGPT &  0.379&0.061&  0.535&0.173&  0.504&0.140&  0.214&0.013&  0.214&0.115&  0.560&0.054&0.401 &0.093\\
         scGPT-sp&  0.605&0.298&  0.582&0.268&  0.571&0.260&  0.407&0.135&  0.666&0.481&  0.640&0.192&0.579 &0.272\\\midrule
         ours(8)&  0.645&0.369&  0.608&0.325&  0.618&0.384&  0.444&0.175&  0.711&0.536&  0.698&0.303&0.621 &0.349\\
         ours(16)
&  0.731&0.504&  0.615&0.340&  0.613&0.367&  0.480&0.217&  0.801&0.725&   \textbf{0.718}&\textbf{0.348}&0.660 &0.417\\
         ours(24)
&  \textbf{0.775}&\textbf{0.595}&  \textbf{0.622}&\textbf{0.360}&  \textbf{0.622}&\textbf{0.399}&  \textbf{0.518}&\textbf{0.260}&  \textbf{0.822}&\textbf{0.767}&  0.710&0.339&\textbf{0.678} &\textbf{0.453}\\
 ours(30)& 0.768&0.578& 0.608&0.341& 0.606&0.352& 0.507&0.244& 0.792&0.726& 0.704&0.324&0.664 &0.428\\ \bottomrule
    \end{tabularx}
\end{table*}

\section{Experiments}
\label{sec:experiments}
\subsection{Datasets}
\noindent\textbf{Pre-training Dataset.} We followed the methodology proposed by scGPT-spatial~\cite{SCGPTSpatial} to construct a pretraining spatial transcriptomics (ST) dataset. This dataset integrates resources from two major public databases, CELLXGENE and GEO, and includes a total of 1,089,785 high-quality spots. It spans multiple systems and organs across embryonic and human samples, including the digestive, respiratory, circulatory, reproductive, nervous, urinary, and endocrine systems. The dataset encompasses common tissue types such as colon, lung, liver, skin, heart, eyes, joints, and muscles, as well as special samples categorized as "others." Moreover, it comprehensively covers a range of physiological and pathological states, including health, inflammation, injury, and various cancer subtypes. This dataset provides abundant spatial gene expression patterns, extensive organ and tissue representativeness, and diverse physiological and pathological state distributions, enabling the model to learn generalized spatial transcriptomic features across tissues and conditions. The rich and diverse data improves the model's adaptability and generalization to complex biological scenarios.

\begin{figure}[t]
\centering
\includegraphics[width=3.0in]{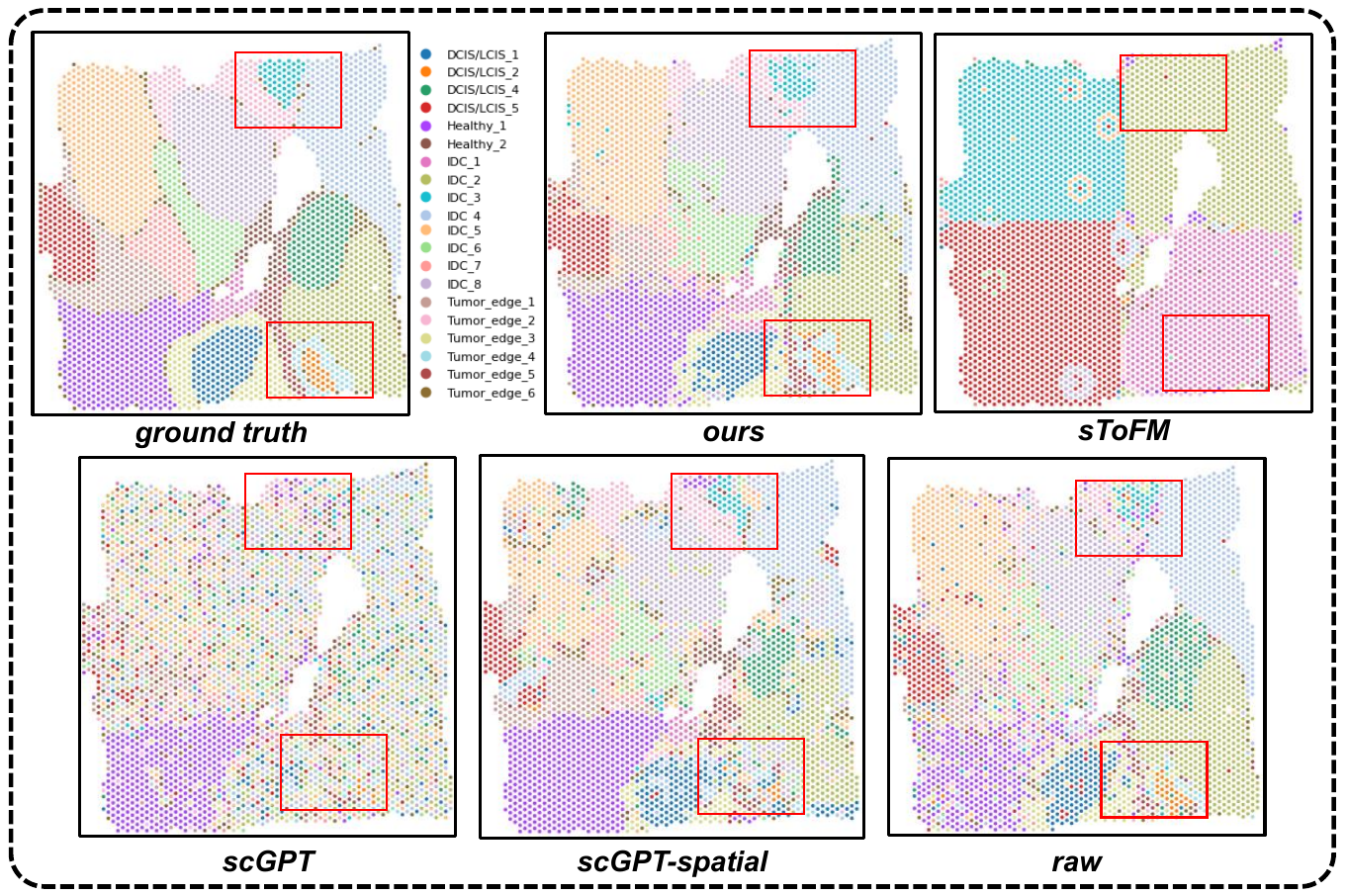}
\caption{Visualization of BrC results (Tables~\ref{tab1} and Tables~\ref{tab:VSsToFM}) — red boxes indicate largest improvements.}
\label{fig-bias}
\end{figure}

\smallskip
\noindent\textbf{Downstream Evaluation Datasets.}
For downstream evaluation, we collect six human ST datasets that provide region-level or cell-type-level ground truth labels.  
These datasets collectively reflect diverse tissue architectures and biological complexity, serving as rigorous benchmarks for assessing model generalization.

Specifically, the \textbf{DLPFC} dataset~\cite{DLPFC} contains about 24,000 spatial spots from human dorsolateral prefrontal cortex samples, each annotated with one of 12 cortical layers.  
The \textbf{Lymph Node A1 (LNA)} and \textbf{Lymph Node D1 (LND)} datasets~\cite{LymphNode} are generated by the 10x Genomics Visium RNA–protein co-assay.  
LNA includes 3,484 spots and 31 proteins categorized into 10 regions, while LND comprises 3,359 spots divided into 11 regions; the two sections differ substantially in cellular composition and are treated as independent datasets.  
The \textbf{Breast Cancer (BrC)} dataset~\cite{EnSDD} consists of 3,798 spots and 36,601 genes annotated into 20 tissue subtypes, providing a challenging benchmark for pathological heterogeneity.  
The \textbf{Tonsil (Ton)} dataset~\cite{SMART_Datasets} captures spatial transcriptomic profiles from three slices of human tonsil tissues, with 13,366 spots classified into four structural domains.  
Finally, the \textbf{Colorectal (Col)} dataset~\cite{Valdeolivas2024} comprises 13 colorectal slices with detailed pathological annotations spanning 17 categories.

All datasets are preprocessed using the same feature space and gene selection methodology as employed during pretraining, ensuring consistency in representation learning and facilitating direct model transfer.

\begin{table*}
    \centering
    \caption{Performance of the proposed method with varying pretraining sample sizes on downstream spatial domain detection tasks using MLP classification. Test accuracy (Acc) and adjusted Rand index (ARI) are reported per dataset and averaged. Size refers to the spatial dimensions (width and height) of pretraining samples.}
    \label{tab3}
    \setlength{\tabcolsep}{5.5pt}
    \setlength{\aboverulesep}{0pt} 
    \setlength{\belowrulesep}{0pt} 
    \begin{tabularx}{\textwidth}{c|cccccccccccccc}\toprule
        \multirow{2}{*}{size}& \multicolumn{2}{c}{DLPFC}& \multicolumn{2}{c}{LNA}& \multicolumn{2}{c}{LND}& \multicolumn{2}{c}{Col}& \multicolumn{2}{c}{BrC}& \multicolumn{2}{c}{Ton}& \multicolumn{2}{c}{Avg}\\\cmidrule{2-15}
 & Acc& ARI & Acc& ARI & Acc& ARI & Acc& ARI & Acc& ARI & Acc& ARI & Acc& ARI\\\midrule
        8 & 0.652&0.370& 0.665  &0.390& 0.609  &0.349& 0.383  &0.143& 0.764  &0.616& 0.659  &0.198& 0.622 &0.344\\
        14 & 0.686&0.440& 0.647  &0.369& 0.614  &0.374& 0.422 &0.157& \textbf{0.780}  &0.679& \textbf{0.680} &0.236& 0.638 &0.376\\
        16 & \textbf{0.753} &\textbf{0.580}& \textbf{0.670} &\textbf{0.393}& \textbf{0.635} &\textbf{0.381}& 0.467 &0.234& 0.773 &\textbf{0.709}
& 0.674 &\textbf{0.268}& \textbf{0.662} &\textbf{0.428}\\
        20 & 0.740  &0.556& 0.664  &0.378& 0.629  &0.354& 0.481 &0.252& 0.706  &0.589& 0.667  &0.216& 0.648 &0.391\\
        24 & 0.720  &0.537& 0.663  &0.390& 0.580  &0.289& 0.491 &0.258& 0.712  &0.629& 0.669  &0.235& 0.639 &0.390\\
        26 & 0.718  &0.535& 0.641  &0.349& 0.586  & 0.288& \textbf{0.493} &\textbf{0.263}& 0.694  &0.586& 0.678  &0.246& 0.635  &0.378\\
        30& 0.667  &0.434& 0.631  &0.348& 0.586  &0.284& 0.482 &0.247& 0.693  &0.604& 0.667  &0.241& 0.621 &0.360\\ \bottomrule
    \end{tabularx}
\end{table*}


\begin{table*}
\centering
\caption{Performance of the proposed method with varying numbers of channels per sample on downstream spatial domain detection tasks using MLP classification. Test accuracy (Acc) and adjusted Rand index (ARI) are reported per dataset and averaged. Channels refers to the number of channels (i.e., selected genes) per pretraining sample.}
\label{tab4}
\setlength{\tabcolsep}{4.75pt}
\setlength{\aboverulesep}{0pt} 
\setlength{\belowrulesep}{0pt} 
\begin{tabularx}{\textwidth}{c|cccccccccccccc}\toprule        
\multirow{2}{*}{channels}& \multicolumn{2}{c}{DLPFC}& \multicolumn{2}{c}{LNA}& \multicolumn{2}{c}{LND}& \multicolumn{2}{c}{Col}& \multicolumn{2}{c}{BrC}& \multicolumn{2}{c}{Ton}& \multicolumn{2}{c}{Avg}\\\cmidrule{2-15}
 & Acc& ARI & Acc& ARI & Acc& ARI & Acc& ARI & Acc& ARI & Acc& ARI & Acc& ARI\\\midrule
512& 0.753 &0.580& \textbf{0.670} &\textbf{0.374}& 0.635 &0.381& 0.467 &0.234& \textbf{0.773} &\textbf{0.709}& \textbf{0.674} &\textbf{0.268}& \textbf{0.662} &\textbf{0.424}\\
256& \textbf{0.766} &\textbf{0.610}& 0.648  &0.334& \textbf{0.640} &\textbf{0.391}& \textbf{0.476} &\textbf{0.244}& 0.759  &0.681& 0.665  &0.240& 0.659  &0.417\\
128& 0.745  &0.570& 0.645  &0.358& 0.620  &0.351& 0.459  &0.231& 0.770  &0.697& 0.663  &0.227& 0.650  &0.406\\ \bottomrule

\end{tabularx}
\end{table*}

\subsection{Implementation Details}
\noindent\textbf{Model Architecture.}
We adopt a masked autoencoder (MAE)-style architecture with a vision transformer (ViT) backbone. 
The encoder comprises six Transformer layers with eight attention heads, a hidden (feed-forward) dimension of 1024, and an embedding dimension of 256. 
The decoder consists of two Transformer layers with identical attention configuration but reduced hidden dimension.  
Each ST sample is represented as an image-like tensor of size $16\times16$ with 512 feature channels corresponding to the selected HVGs.  
Spatial coordinates within each tissue slice are min–max normalized to the range [0, 100] to maintain compact non-overlapping layouts.

\smallskip
\noindent\textbf{Pretraining Configuration.}
We train the model for 40 epochs using the Adam optimizer (learning rate $1\times10^{-4}$, $\beta_1=0.9$, $\beta_2=0.999$, weight decay=0.01) and a cosine learning rate scheduler.  
The batch size is set to 32, and the masking ratio is 30\%.  
All pretraining experiments are conducted on eight GeForce RTX 4090 GPUs (24GB each), with a total training time of approximately 32 GPU-nakow.  

\smallskip
\noindent\textbf{Baselines.}
We compare against:
\begin{itemize}
    \item \textbf{scGPT}~\cite{cui2023scGPT}: transformer trained on single-cell data with spot-level sampling.
    \item \textbf{scGPT-spatial}~\cite{SCGPTSpatial}: incorporates 16 nearest neighbors for spatial context.
    \item \textbf{Raw Expression~\cite{raw-classification}}: directly uses expression vectors.
\end{itemize}
All baselines use identical Transformer depth, hidden dimensions, and input preprocessing for fair comparison.

\smallskip
\noindent\textbf{Downstream Evaluation.}
For Spatial Domain Detection, we freeze the pretrained weights and train only lightweight classifiers to avoid feature leakage. A strict 20\% training and 80\% testing split is used to evaluate few-shot generalization.
Two classifier heads are considered:
(1){MLP classifier:} 3-layer MLP with hidden sizes [512, 256, 128], trained using Adam ($\mathrm{lr}=1\times10^{-3}$, batch size=32) for 800 epochs. 
(2){k-NN classifier:} Non-parametric classifier with $k=10$ and Euclidean distance. Each dataset is evaluated over 10 random splits, and the mean accuracy/ARI is reported.

For Masked Region Reconstruction, we mask a contiguous $S \times S$ region and reconstruct the top 512 HVGs. 
Performance is evaluated using MSE and MAE between predicted and ground-truth expression matrices. 
Our method uses all unmasked spots within a bounding region for contextual inference, while scGPT-spatial averages embeddings from the 16 nearest neighbors. For each experiment, we did twenty replications and averaged the results.

\begin{table*}
\centering
\caption{Performance comparison of different channel selection methods for downstream spatial domain detection using MLP classification. Test accuracy (Acc) and adjusted Rand index (ARI) are reported per dataset and averaged. Weighted: proposed weighted selection; HVGs: highly variable genes; Random: random channel selection.}
\label{tab5}
\setlength{\tabcolsep}{4.8pt}
\setlength{\aboverulesep}{0pt} 
\setlength{\belowrulesep}{0pt} 
\begin{tabularx}{\textwidth}{c|cccccccccccccc}\toprule        
\multirow{2}{*}{method} & \multicolumn{2}{c}{DLPFC}& \multicolumn{2}{c}{LNA}& \multicolumn{2}{c}{LND}& \multicolumn{2}{c}{Col}& \multicolumn{2}{c}{BrC}& \multicolumn{2}{c}{Ton}& \multicolumn{2}{c}{Avg}\\\cmidrule{2-15}
 & Acc& ARI & Acc& ARI & Acc& ARI & Acc& ARI & Acc& ARI & Acc& ARI & Acc& ARI\\\midrule
Weighted& \textbf{0.753} &\textbf{0.580}& \textbf{0.670} &\textbf{0.393}& \textbf{0.640} &\textbf{0.381}& \textbf{0.467} &\textbf{0.234}& 0.773 &0.709& 0.674 &0.268& \textbf{0.663} &\textbf{0.428}\\
HVGs& 0.725 &0.536& 0.659 &0.357& 0.619 &0.347& 0.433 &0.220& \textbf{0.788} &\textbf{0.718}& \textbf{0.688} &\textbf{0.281}& 0.652 &0.410\\
Random& 0.688 &0.466& 0.647 &0.341& 0.635 &0.360& 0.457 & 0.155& 0.749 &0.684& 0.646 & 0.163& 0.637 &0.362\\ \bottomrule

\end{tabularx}

\end{table*}

\begin{table*}
    \centering
    \caption{Masked region reconstruction performance on DLPFC data, measured by mean squared error (MSE) and mean absolute error (MAE) for our method and the scGPT-spatial baseline across different masking ratios. Note: scGPT and raw methods cannot perform this task due to lack of spatial information. Size refers to the spatial
dimensions (width and height) of pretraining samples in our method.}
    \label{tab6}
    \setlength{\aboverulesep}{0pt} 
    \setlength{\belowrulesep}{0pt} 
    \begin{tabular}{c|c|cccccccccccc}\toprule
         \multirow{2}{*}{method} & \multirow{2}{*}{size} & \multicolumn{2}{c}{4} &  \multicolumn{2}{c}{8} &  \multicolumn{2}{c}{10}& \multicolumn{2}{c}{14} & \multicolumn{2}{c}{16}& \multicolumn{2}{c}{18}\\\cmidrule{3-14}
         & & MSE& MAE & MSE&MAE & MSE&MAE & MSE& MAE  & MSE&MAE  & MSE&MAE \\\midrule
         \multirow{5}{*}{ours}& 30& 0.251&  0.190 &  \textbf{0.229}&  0.187& \textbf{0.241}& \textbf{0.185}& \textbf{0.243}& \textbf{0.187}  & 0.250&0.188& 0.251&0.190 \\
 &26& 0.251& 0.195 & 0.250& 0.194 & 0.254&0.197& 0.256& 0.196  & 0.246&0.192& \textbf{0.237}&0.188\\
 &24& \textbf{0.243}& \textbf{0.187} & 0.253& 0.190 & 0.253&0.189& 0.256& 0.191  & 0.249&0.189& 0.247&\textbf{0.186}\\
 & 22& 0.252& 0.189& 0.237& \textbf{0.185}& 0.244& 0.188& 0.252& 0.192 & \textbf{0.231}&\textbf{0.186}& 0.254& 0.190\\
 & 20& 0.234& 0.205& 0.244& 0.207& 0.246& 0.190& 0.246& 0.213 & 0.238&0.210& 0.254&  0.219\\
         scGPT-sp&-&0.333&  0.333 &  0.329&  0.334 & 0.340&0.339& 0.320& 0.328& 0.341&0.342& 0.326&0.332\\ \bottomrule
    \end{tabular}
\end{table*}

\subsection{Results and Analysis}

We systematically evaluate our method by analyzing experimental results in a question-and-answer (QA) format, focusing on performance comparison, robustness, and hyperparameter sensitivity.

\noindent\textit{Q1:}\textit{How does the proposed sampling strategy outperform existing methods?}

\noindent\textit{A1:}Notably, the red box in Figure~\ref{fig-bias} most clearly illustrates the improvements achieved by our method. Furthermore, as shown in Tables~\ref{tab1} and~\ref{tab2}, models trained with our sampling strategy consistently attain the best performance across nearly all datasets in the Spatial Domain Detection task.

\begin{itemize}
    \item Comparison with spot-based pretraining:  
    Compared with scGPT~\cite{cui2023scGPT}, which only employs spot-level sampling, our approach achieves substantial average improvements of 0.287 in Acc and 0.347 in ARI, and 0.277 in Acc and 0.360 in ARI.  
    Remarkably, on the breast cancer dataset, the performance gains reach as high as 0.568 in Acc and 0.599 in ARI, and 0.608 in Acc and 0.652 in ARI.  
    These results demonstrate that our sample construction strategy more effectively preserves the intrinsic spatial organization embedded in spatial transcriptomics (ST) data.

    \item Comparison with multi-spot pretraining:  
    Even when compared with scGPT-spatial~\cite{SCGPTSpatial}—which integrates neighboring gene expressions as spatial compensation—our method still achieves consistent performance advantages, with average gains of 0.059 in Acc and 0.086 in ARI, and 0.099 in Acc and 0.181 in ARI.  
    This further verifies the superior capacity of our sampling strategy to capture spatial dependencies and structural context.
\end{itemize}

In the Masked Region Reconstruction task (as shown in Table~\ref{tab6}), our method also surpasses scGPT-spatial~\cite{SCGPTSpatial} across all mask sizes, consistently yielding lower MSE and MAE values—by approximately 0.09 and 0.15 on average.  
These results confirm that our sample construction strategy enables the model to learn more comprehensive spatial and transcriptional representations, thereby improving generalization and enhancing downstream performance such as gene expression recovery.



\noindent\textit{Q2:}\textit{How robust is the proposed sample construction strategy across diverse datasets?}

\noindent\textit{A2:}As shown in Tables~\ref{tab1}, Table~\ref{tab2}, and Table~\ref{tab6}, our method consistently outperforms all baseline approaches across nearly all datasets in both the Spatial Domain Detection and Masked Region Reconstruction tasks.  
These datasets encompass spatially resolved transcriptomics (ST) data derived from a wide range of human tissues, each characterized by substantial heterogeneity in cellular composition, spatial architecture, and biological function.

Such consistent improvements across structurally and distributionally diverse datasets highlight the remarkable robustness and strong generalization capacity of our proposed sampling strategy.  
By adaptively capturing spatial dependencies and transcriptomic variation, our method maintains stable performance even under substantial shifts in tissue type and spatial complexity.  
This indicates that the strategy not only generalizes well to unseen biological contexts but also provides a reliable foundation for pretraining large-scale ST models across heterogeneous data sources.

\noindent\textit{Q3:}\textit{Does sample size affect pre-training quality?} 


\noindent\textit{A3:} As shown in Table~\ref{tab3}, the sample size substantially affects the quality of pretraining, but the relationship is non-monotonic. Our results reveal an optimal range that balances spatial context richness and noise control.

Moderate enlargement is beneficial: Expanding the sample size from $8\times8$ to $16\times16$ provides a broader spatial context, enabling the model to capture more coherent local dependencies. This leads to a clear improvement in performance, with average accuracy (Acc) rising from 0.622 to 0.662 and adjusted Rand index (ARI) from 0.344 to 0.428. This observation echoes the scaling law principle: increasing data coverage enhances representation learning.
     
Excessive size is detrimental: Further expanding the window to $30\times30$ results in a decline of 0.041 in Acc and 0.068 in ARI relative to the $16\times16$ setting. Oversized samples introduce redundant and heterogeneous background information, which weakens the model’s ability to capture precise local spatial dependencies. This aligns with findings that excessively large data can introduce noise and lead to overfitting or performance saturation.

\noindent\textit{Q4}:\textit{Does channel count affect pre-training quality?} 

\noindent\textit{A4:}As shown in Table~\ref{tab4}, the number of channels (i.e., selected genes) substantially influences pretraining quality, but its effect is dataset-dependent. Increasing channels generally improves average performance, yet beyond a certain point, more channels do not guarantee better results. While using 512 channels achieve the highest overall average, 256 channels outperform 512 on DLPFC, LND and Col dataset, and 128 channels exceed 256 by 0.011 on Acc and 0.016 on ARI on the BrC dataset. This indicates that matching representational capacity to dataset variability is critical. A 512-channel setting serves as a robust general default, but selective reduction can yield gains on specific cohorts. 

\noindent\textit{Q5:}\textit{In downstream tasks, do different channel selections make a difference?}

\noindent\textit{A5:} The choice of gene selection strategy affects results, and the best method depends on the dataset. While adaptive methods work well, weighted random sampling is usually the most reliable across different datasets. As shown in Table~\ref{tab5}, the weighted random sampling approach—which mirrors the strategy used during pretraining—achieves an average performance gain of 0.011 on Acc and 0.018 on ARI over selecting only highly variable genes, and 0.026 on Acc 0.062 on ARI over pure random selection. It delivers the best results on the majority of evaluated datasets. But on the BrC and Ton datasets, directly selecting highly variable genes yields better performance, with improvements of 0.015 on Acc and 0.009 on ARI over the weighted random sampling approach. 

\noindent\textit{Q6:}\textit{What is the recommended configuration of the pretraining sample?}

 \noindent\textit{A6:} Across all experiments, the 16×16×512 configuration provides the most reliable overall performance. As shown in Table~\ref{tab1}, this setting achieves the best average performance (0.662 on Acc, 0.428 on ARI) among all candidates. In Table~\ref{tab2}, it also
delivers highly competitive results, trailing the best-performing configuration by only 0.018 on Acc, 0.036 on ARI.These findings indicate that a 16×16 spatial window combined with 512 gene channels offers a strong balance between spatial context, representational capacity, and computational efficiency, making it a robust default choice for pretraining.

\noindent\textit{Q7:}\textit{By constructing ST samples as multi‑channel images, can we enable the effective application of computer vision(CV) techniques to spatial transcriptomics data?}

 \noindent\textit{A7:} In our experiments, the model architecture and training method adopt the visual transformer (ViT~\cite {vit}) and masking autoencoder (MAE~\cite{MAE}) frameworks in the field of computer vision, respectively. This demonstrates that advanced computer vision technologies can be effectively leveraged for large-scale spatial transcriptomics by our sampling strategy, providing a robust foundation for developing more powerful ST models. 

\noindent\textit{Q8:}\textit{Why does the baseline exclude other graph-based spatial methods or slice-level methods?}

 \noindent\textit{A8:} Existing graph-based methods (e.g., GraphST~\cite{GraphST}, STAGATE~\cite{STAGATE}) are task-specific spatial-transcriptomics approaches, not foundation models. And slice-level methods (such as sToFM~\cite{stofm}) have substantially different architectures from our model — which uses the scGPT stacked-transformer encoder–decoder. To isolate the effect of sample construction we exclude them from our baseline. We still compared to sToFM~\cite{stofm} using its supplied weights (no training code available). Table~\ref{tab:VSsToFM} reports a comparison between our model (trained on 1M spots) and the supplied sToFM model (trained on 88M spots): 0.662 vs. 0.542. 
 
 \noindent\textit{Q9:Will this sample construction distort the original physical distance and destroy the original organization?}
 \noindent\textit{A9:} Figure~\ref{fig-samples} shows the process of processing a slice into patches, where the relative positions of individual spots remain unchanged. Additionally, all datasets use the 10x Visium platform (uniform resolution), the uniform operation of all data does not distort the physical distance and organizational structure.

\section{Conclusion}
\label{sec:Conclusion}

Spatial transcriptomics (ST) pretraining struggles to balance spatial context preservation and computational scalability, as current spot- and slice-based approaches are limited. We address this by treating ST data as croppable multi-channel images, extracting fixed-size patches with importance-aware gene selection. This approach preserves local structure, boosts sample diversity, and enables scalable pretraining. Experiments show our method consistently outperforms existing ones across tasks and datasets. By standardizing ST data and enabling a flexible pretraining pipeline, we establish a strong foundation for spatial models and facilitate the application of computer vision techniques to spatial biology.

\section{Acknowledgments}
This work was supported by the Strategic Priority Research Program of the Chinese Academy of Sciences under Grant No. XDA0460205. We thank Shopee for providing computing power support.

{
    \small
    \bibliographystyle{ieeenat_fullname}
    \bibliography{main}

@String(CVPR= {IEEE Conf. Comput. Vis. Pattern Recog.})

@String(CVPRW= {IEEE Conf. Comput. Vis. Pattern Recog. Worksh.})

@String(CVPR  = {CVPR})

@String(CVPRW= {CVPRW})

@article{Systematic_comparison, title={Systematic comparison of sequencing-based spatial transcriptomic methods}, volume={21}, url={http://dx.doi.org/10.1038/s41592-024-02325-3}, DOI={10.1038/s41592-024-02325-3}, number={9}, journal={Nature Methods}, publisher={Springer Science and Business Media LLC}, author={You, Yue and Fu, Yuting and Li, Lanxiang and Zhang, Zhongmin and Jia, Shikai and Lu, Shihong and Ren, Wenle and Liu, Yifang and Xu, Yang and Liu, Xiaojing and Jiang, Fuqing and Peng, Guangdun and Sampath Kumar, Abhishek and Ritchie, Matthew E. and Liu, Xiaodong and Tian, Luyi}, year={2024}, month=july, pages={1743–1754}, language={en} }

@article{raw-classification, title={Joint cell segmentation and cell type annotation for spatial transcriptomics}, volume={17}, url={http://dx.doi.org/10.15252/msb.202010108}, DOI={10.15252/msb.202010108}, number={6}, journal={Molecular Systems Biology}, publisher={Springer Science and Business Media LLC}, author={Littman, Russell and Hemminger, Zachary and Foreman, Robert and Arneson, Douglas and Zhang, Guanglin and Gómez‐Pinilla, Fernando and Yang, Xia and Wollman, Roy}, year={2021}, month=may, language={en} }

@article{Marx_2021, title={Method of the Year: spatially resolved transcriptomics}, volume={18}, url={http://dx.doi.org/10.1038/s41592-020-01033-y}, DOI={10.1038/s41592-020-01033-y}, number={1}, journal={Nature Methods}, publisher={Springer Science and Business Media LLC}, author={Marx, Vivien}, year={2021}, month=jan, pages={9–14}, language={en} }

@article{SpatialScope, title={Integrating spatial and single-cell transcriptomics data using deep generative models with SpatialScope}, volume={14}, url={http://dx.doi.org/10.1038/s41467-023-43629-w}, DOI={10.1038/s41467-023-43629-w}, number={1}, journal={Nature Communications}, publisher={Springer Science and Business Media LLC}, author={Wan, Xiaomeng and Xiao, Jiashun and Tam, Sindy Sing Ting and Cai, Mingxuan and Sugimura, Ryohichi and Wang, Yang and Wan, Xiang and Lin, Zhixiang and Wu, Angela Ruohao and Yang, Can}, year={2023}, month=nov, language={en} }

@article{Burgess_2019, title={Spatial transcriptomics coming of age}, volume={20}, url={http://dx.doi.org/10.1038/s41576-019-0129-z}, DOI={10.1038/s41576-019-0129-z}, number={6}, journal={Nature Reviews Genetics}, publisher={Springer Science and Business Media LLC}, author={Burgess, Darren J.}, year={2019}, month=apr, pages={317–317}, language={en} }

@article{YANG2024109966,
title = {Spatial transcriptomics analysis of gene expression prediction using exemplar guided graph neural network},
journal = {Pattern Recognition},
volume = {145},
pages = {109966},
year = {2024},
issn = {0031-3203},
doi = {https://doi.org/10.1016/j.patcog.2023.109966},
url = {https://www.sciencedirect.com/science/article/pii/S0031320323006647},
author = {Yan Yang and Md Zakir Hossain and Eric Stone and Shafin Rahman},
}

@article{Asp_2020,
author = {Asp, Michaela and Bergenstråhle, Joseph and Lundeberg, Joakim},
year = {2020},
month = {05},
pages = {1900221},
title = {Spatially Resolved Transcriptomes—Next Generation Tools for Tissue Exploration},
volume = {42},
journal = {BioEssays},
doi = {10.1002/bies.201900221}
}

@article{Miller2021MultiSD,
  title={Multi scale diffeomorphic metric mapping of spatial transcriptomics datasets},
  author={Michael I. Miller and Jean Fan and Daniel Jacob Tward},
  journal={2021 IEEE/CVF Conference on Computer Vision and Pattern Recognition Workshops (CVPRW)},
  year={2021},
  pages={4467-4475},
  url={https://api.semanticscholar.org/CorpusID:235692272}
}

@article{Chung2024AccurateSG,
  title={Accurate Spatial Gene Expression Prediction by Integrating Multi-Resolution Features},
  author={Youngmin Chung and Ji Hun Ha and Kyeong Chan Im and Joo Sang Lee},
  journal={2024 IEEE/CVF Conference on Computer Vision and Pattern Recognition (CVPR)},
  year={2024},
  pages={11591-11600},
  url={https://api.semanticscholar.org/CorpusID:268363616}
}

@article{SCGPTSpatial,
  title={scGPT-spatial: Continual Pretraining of Single-Cell Foundation Model for Spatial Transcriptomics},
  author={Wang, Chloe Xueqi and Cui, Haotian and Zhang, Andrew Hanzhuo and Xie, Ronald and Goodarzi, Hani and Wang, Bo},
  journal={bioRxiv},
  pages={2025--02},
  year={2025},
  publisher={Cold Spring Harbor Laboratory}
}

@article{cui2023scGPT,
title={scGPT: Towards Building a Foundation Model for Single-Cell Multi-omics Using Generative AI},
author={Cui, Haotian and Wang, Chloe and Maan, Hassaan and Pang, Kuan and Luo, Fengning and Wang, Bo},
journal={bioRxiv},
year={2023},
publisher={Cold Spring Harbor Laboratory}
}

@article{
Visium,
author = {Patrik L. Ståhl  and Fredrik Salmén  and Sanja Vickovic  and Anna Lundmark  and José Fernández Navarro  and Jens Magnusson  and Stefania Giacomello  and Michaela Asp  and Jakub O. Westholm  and Mikael Huss  and Annelie Mollbrink  and Sten Linnarsson  and Simone Codeluppi  and Åke Borg  and Fredrik Pontén  and Paul Igor Costea  and Pelin Sahlén  and Jan Mulder  and Olaf Bergmann  and Joakim Lundeberg  and Jonas Frisén },
title = {Visualization and analysis of gene expression in tissue sections by spatial transcriptomics},
journal = {Science},
volume = {353},
number = {6294},
pages = {78-82},
year = {2016},
doi = {10.1126/science.aaf2403},
URL = {https://www.science.org/doi/abs/10.1126/science.aaf2403},
eprint = {https://www.science.org/doi/pdf/10.1126/science.aaf2403},
}

@article{Xenium,
       author = {{Janesick}, Amanda and {Shelansky}, Robert and {Gottscho}, Andrew D. and {Wagner}, Florian and {Williams}, Stephen R. and {Rouault}, Morgane and {Beliakoff}, Ghezal and {Morrison}, Carolyn A. and {Oliveira}, Michelli F. and {Sicherman}, Jordan T. and {Kohlway}, Andrew and {Abousoud}, Jawad and {Drennon}, Tingsheng Yu and {Mohabbat}, Seayar H. and {10x Development Teams} and {Taylor}, Sarah E.~B.},
        title = "{High resolution mapping of the tumor microenvironment using integrated single-cell, spatial and in situ analysis}",
      journal = {Nature Communications},
         year = 2023,
        month = dec,
       volume = {14},
          eid = {8353},
        pages = {8353},
          doi = {10.1038/s41467-023-43458-x},
       adsurl = {https://ui.adsabs.harvard.edu/abs/2023NatCo..14.8353J}
}

@article{
MERFISH,
author = {Kok Hao Chen  and Alistair N. Boettiger  and Jeffrey R. Moffitt  and Siyuan Wang  and Xiaowei Zhuang },
title = {Spatially resolved, highly multiplexed RNA profiling in single cells},
journal = {Science},
volume = {348},
number = {6233},
pages = {aaa6090},
year = {2015},
doi = {10.1126/science.aaa6090},
URL = {https://www.science.org/doi/abs/10.1126/science.aaa6090},
eprint = {https://www.science.org/doi/pdf/10.1126/science.aaa6090},
}

@article{He2020IntegratingSG,
  title={Integrating spatial gene expression and breast tumour morphology via deep learning},
  author={Bryan He and Ludvig Bergenstr{\aa}hle and Linnea Stenbeck and Abubakar Abid and Alma Andersson and {\AA}ke Borg and Jonas Maaskola and Joakim Lundeberg and James Y. Zou},
  journal={Nature Biomedical Engineering},
  year={2020},
  volume={4},
  pages={827 - 834},
  url={https://api.semanticscholar.org/CorpusID:219977369}
}

@article{STAGATE, title={Deciphering spatial domains from spatially resolved transcriptomics with an adaptive graph attention auto-encoder}, volume={13}, url={http://dx.doi.org/10.1038/s41467-022-29439-6}, DOI={10.1038/s41467-022-29439-6}, number={1}, journal={Nature Communications}, publisher={Springer Science and Business Media LLC}, author={Dong, Kangning and Zhang, Shihua}, year={2022}, month=apr, language={en} }

@article{GraphST, title={Spatially informed clustering, integration, and deconvolution of spatial transcriptomics with GraphST}, volume={14}, url={http://dx.doi.org/10.1038/s41467-023-36796-3}, DOI={10.1038/s41467-023-36796-3}, number={1}, journal={Nature Communications}, publisher={Springer Science and Business Media LLC}, author={Long, Yahui and Ang, Kok Siong and Li, Mengwei and Chong, Kian Long Kelvin and Sethi, Raman and Zhong, Chengwei and Xu, Hang and Ong, Zhiwei and Sachaphibulkij, Karishma and Chen, Ao and Zeng, Li and Fu, Huazhu and Wu, Min and Lim, Lina Hsiu Kim and Liu, Longqi and Chen, Jinmiao}, year={2023}, month=mar, language={en}}

@article{STitch3D, title={Construction of a 3D whole organism spatial atlas by joint modelling of multiple slices with deep neural networks}, volume={5}, url={http://dx.doi.org/10.1038/s42256-023-00734-1}, DOI={10.1038/s42256-023-00734-1}, number={11}, journal={Nature Machine Intelligence}, publisher={Springer Science and Business Media LLC}, author={Wang, Gefei and Zhao, Jia and Yan, Yan and Wang, Yang and Wu, Angela Ruohao and Yang, Can}, year={2023}, month=oct, pages={1200–1213}, language={en} }

@article{
doi:10.1126/science.aav9776,
author = {Silas Maniatis  and Tarmo Äijö  and Sanja Vickovic  and Catherine Braine  and Kristy Kang  and Annelie Mollbrink  and Delphine Fagegaltier  and Žaneta Andrusivová  and Sami Saarenpää  and Gonzalo Saiz-Castro  and Miguel Cuevas  and Aaron Watters  and Joakim Lundeberg  and Richard Bonneau  and Hemali Phatnani },
title = {Spatiotemporal dynamics of molecular pathology in amyotrophic lateral sclerosis},
journal = {Science},
volume = {364},
number = {6435},
pages = {89-93},
year = {2019},
doi = {10.1126/science.aav9776},
URL = {https://www.science.org/doi/abs/10.1126/science.aav9776},
eprint = {https://www.science.org/doi/pdf/10.1126/science.aav9776},
}

@article{2019A,
  title={A Spatiotemporal Organ-Wide Gene Expression and Cell Atlas of the Developing Human Heart},
  author={ Asp, Michaela  and  Giacomello, Stefania  and  Larsson, Ludvig  and  Wu, Chenglin  and Daniel Fürth and  Qian, Xiaoyan  and  Wrdell, Eva  and  Custodio, Joaquin  and  Reimegrd, Johan  and Fredrik Salmén},
  journal={Cell},
  volume={179},
  number={7},
  pages={1647-1660.e19},
  year={2019},
}

@article{
MERFISH2,
author = {Jeffrey R. Moffitt  and Dhananjay Bambah-Mukku  and Stephen W. Eichhorn  and Eric Vaughn  and Karthik Shekhar  and Julio D. Perez  and Nimrod D. Rubinstein  and Junjie Hao  and Aviv Regev  and Catherine Dulac  and Xiaowei Zhuang },
title = {Molecular, spatial, and functional single-cell profiling of the hypothalamic preoptic region},
journal = {Science},
volume = {362},
number = {6416},
pages = {eaau5324},
year = {2018},
doi = {10.1126/science.aau5324},
URL = {https://www.science.org/doi/abs/10.1126/science.aau5324},
eprint = {https://www.science.org/doi/pdf/10.1126/science.aau5324},
}

@article{Lin2024STAlignAM,
  title={ST-Align: A Multimodal Foundation Model for Image-Gene Alignment in Spatial Transcriptomics},
  author={Yuxiang Lin and Ling Luo and Ying Chen and Xushi Zhang and Zihui Wang and Wenxian Yang and Mengsha Tong and Rongshan Yu},
  journal={ArXiv},
  year={2024},
  volume={abs/2411.16793},
  url={https://api.semanticscholar.org/CorpusID:274280682}
}

@article{stofm,
  title={SToFM: a Multi-scale Foundation Model for Spatial Transcriptomics},
  author={Zhao, Suyuan and Luo, Yizhen and Yang, Ganbo and Zhong, Yan and Zhou, Hao and Nie, Zaiqing},
  journal={arXiv preprint arXiv:2507.11588},
  year={2025}
}

@article{FmH2ST,
    author = {Wang, Yuequn and Wang, Jun and Xu, Yanyu and Liu, Ning and Liu, Bin and Li, Yuliang and Yu, Guoxian},
    title = {FmH2ST: foundation model-based spatial transcriptomics generation from histological images},
    journal = {Nucleic Acids Research},
    volume = {53},
    number = {17},
    pages = {gkaf865},
    year = {2025},
    month = {09},
    issn = {1362-4962},
    doi = {10.1093/nar/gkaf865},
    url = {https://doi.org/10.1093/nar/gkaf865},
    eprint = {https://academic.oup.com/nar/article-pdf/53/17/gkaf865/64870080/gkaf865.pdf},
}

@article{DLPFC,
  title={Transcriptome-scale spatial gene expression in the human dorsolateral prefrontal cortex},
  author={Maynard, Kristen R and Collado-Torres, Leonardo and Weber, Lukas M and Uytingco, Cedric and Barry, Brianna K and Williams, Stephen R and Catallini, Joseph L and Tran, Matthew N and Besich, Zachary and Tippani, Madhavi and others},
  journal={Nature neuroscience},
  volume={24},
  number={3},
  pages={425--436},
  year={2021},
  publisher={Nature Publishing Group US New York}
}

@article{LymphNode,
title={Deciphering spatial domains from spatial multi-omics with SpatialGlue},
volume={21},
ISSN={1548-7105},
url={http://dx.doi.org/10.1038/s41592-024-02316-4},
DOI={10.1038/s41592-024-02316-4},
number={9},
journal={Nature Methods},
publisher={Springer Science and Business Media LLC}, author={Long, Yahui and Ang, Kok Siong and Sethi, Raman and Liao, Sha and Heng, Yang and van Olst, Lynn and Ye, Shuchen and Zhong, Chengwei and Xu, Hang and Zhang, Di and Kwok, Immanuel and Husna, Nazihah and Jian, Min and Ng, Lai Guan and Chen, Ao and Gascoigne, Nicholas R. J. and Gate, David and Fan, Rong and Xu, Xun and Chen, Jinmiao}, year={2024}, month=jun, pages={1658-1667} }

@article{EnSDD,
title={Enhancing spatial domain detection in spatial transcriptomics with EnSDD},
volume={7}, ISSN={2399-3642}, url={http://dx.doi.org/10.1038/s42003-024-07001-y},
DOI={10.1038/s42003-024-07001-y},
number={1},
journal={Communications Biology},
publisher={Springer Science and Business Media LLC}, author={Li, Hui-Sheng and Tan, Yu-Ting and Zhang, Xiao-Fei}, year={2024}, month=oct }

@misc{SMART_Datasets,
  author    = {Zheng, X.},
  title     = {Datasets used in SMART’s experiments},
  howpublished = {Zenodo dataset, DOI:10.5281/zenodo.17093158},
  year      = {2025},
  url       = {https://doi.org/10.5281/zenodo.17093158}
}

@article{Valdeolivas2024,
  author    = {Valdeolivas, Alberto and Amberg, Bettina and Giroud, Nicolas and Richardson, Marion and Gálvez, Eric J. C. and Badillo, Solveig and Julien-Laferrière, Alice and Túrós, Demeter and Voith von Voithenberg, Lena and Wells, Isabelle and Pesti, Benedek and Lo, Amy A. and Yángüez, Emilio and Das Thakur, Meghna and Bscheider, Michael and Sultan, Marc and Kumpesa, Nadine and Jacobsen, Björn and Bergauer, Tobias and Saez-Rodriguez, Julio and Rottenberg, Sven and Schwalie, Petra C. and Hahn, Kerstin},
  title     = {Profiling the heterogeneity of colorectal cancer consensus molecular subtypes using spatial transcriptomics},
  journal   = {npj Precision Oncology},
  volume    = {8},
  number    = {1},
  pages     = {10},
  year      = {2024},
  doi       = {10.1038/s41698-023-00488-4},
  url       = {http://dx.doi.org/10.1038/s41698-023-00488-4}
}

@article{MAE,
  title={Masked Autoencoders Are Scalable Vision Learners},
  author={Kaiming He and Xinlei Chen and Saining Xie and Yanghao Li and Piotr Doll'ar and Ross B. Girshick},
  journal={2022 IEEE/CVF Conference on Computer Vision and Pattern Recognition (CVPR)},
  year={2021},
  pages={15979-15988},
  url={https://api.semanticscholar.org/CorpusID:243985980}
}

@article{Vit,
  title={An Image is Worth 16x16 Words: Transformers for Image Recognition at Scale},
  author={Alexey Dosovitskiy and Lucas Beyer and Alexander Kolesnikov and Dirk Weissenborn and Xiaohua Zhai and Thomas Unterthiner and Mostafa Dehghani and Matthias Minderer and Georg Heigold and Sylvain Gelly and Jakob Uszkoreit and Neil Houlsby},
  journal={ArXiv},
  year={2020},
  volume={abs/2010.11929},
  url={https://api.semanticscholar.org/CorpusID:225039882}
}
}



\end{document}